# HeFS: Helper-Enhanced Feature Selection via Pareto-Optimized Genetic Search

Yusi Fan [1,2,*], Tian Wang [3,*], Zhiying Yan [4,*], Chang Liu [5], Qiong Zhou [1,2], Qi Lu [1,2], Zhehao Guo [1,2], Ziqi Deng [6], Wenyu Zhu [4,#], Ruochi Zhang [1,2,#], Fengfeng Zhou [1,2,#].

1 College of Computer Science and Technology, Jilin University, Changchun, China, 130012.

2 Key Laboratory of Symbolic Computation and Knowledge Engineering of Ministry of Education, Jilin University, Changchun, China, 130012.

3 Department of Chemical and Biological Engineering, The Hong Kong University of Science and Technology, Clear Water Bay, Hong Kong, China, 999077

4 Department of Oncology, The Second People's Hospital of Changzhou, the Third Affiliated Hospital of Nanjing Medical University, Changzhou, China, 213003.

5 Beijing Life Science Academy, Beijing, China, 102209

6 School of Science, The Hong Kong University of Science and Technology, Hong Kong, China, 999077

* These authors contribute equally to this study.

# Correspondence may be addressed to Fengfeng Zhou: FengfengZhou@gmail.com or ffzhou@jlu.edu.cn . Lab web site: http://www.healthinformaticslab.org/ . Phone: +86-431-8516-6024. Fax: +86-431-8516-6024. Correspondence may also be addressed to Ruochi Zhang: zrc720@gmail.com and Wenyu Zhu: wenyu.zhu@njmu.edu.cn.

## Abstract

Feature selection is a combinatorial optimization problem that is NP-hard. Conventional approaches often employ heuristic or greedy strategies, which are prone to premature convergence and may fail to capture subtle yet informative features. This limitation becomes especially critical in high-dimensional datasets, where complex and interdependent feature relationships prevail. We introduce the HeFS (Helper-Enhanced Feature Selection) framework to refine feature subsets produced by existing algorithms. HeFS systematically searches the residual feature space to identify a *Helper Set*—features that complement the original subset and improve classification performance. The approach employs a biased initialization scheme and a ratio-guided mutation mechanism within a genetic algorithm, coupled with Pareto-based multi-objective optimization to jointly maximize predictive accuracy and feature complementarity. Experiments on 18 benchmark datasets demonstrate that HeFS consistently identifies overlooked yet informative features and achieves superior performance over state-of-the-art methods, including in challenging domains such as gastric cancer classification, drug toxicity prediction, and computer science applications. The code and datasets are available at https://healthinformaticslab.org/supp/.

## 1 Introduction

With the rapid development of machine learning and deep learning algorithms for data-driven prediction tasks, feature selection has become an essential step in building robust models. Its primary goal is to select a subset of informative features from a large initial set to improve predictive performance and interpretability (Qian, et al., 2023; Rostami, et al., 2022). This step is especially critical for high-dimensional settings, such as chemical molecular datasets (Yusof, Muda and Pratama, 2021), which can contain thousands of features.

Conventional univariate feature selection methods face significant challenges in such settings, where strong interdependencies between features can obscure their individual statistical significance. Features that appear weak in isolation may nevertheless provide substantial predictive value when combined with others. Furthermore, the combinatorial nature of the problem yields an immense search space of size $2^n$ for the *n* original features (Ahadzadeh, et al., 2023), and makes exhaustive evaluation infeasible and rendering the problem NP-hard (Wei, et al., 2023; Xue, Zhang and Browne, 2012).

Feature selection can be viewed as a search problem (Abdel-Basset, et al., 2020; Wang, Xiao and Rajasekaran, 2020) (Han, Huang and Zhou, 2021), solvable via exhaustive, random, or heuristic strategies (Abdulwahab, et al., 2024; Wang, Wang and Chang, 2016). Exhaustive search evaluates all feature combinations, guaranteeing optimality but suffering from exponential complexity, which is prohibitive for a large number of features. Random search (Bischl, et al., 2023) reduces computational cost but lacks systematic exploration, often producing suboptimal results in high-dimensional spaces.

Heuristic search methods leverage problem-specific strategies to efficiently traverse the search space, including simulated annealing(Shi, et al., 2023), particle swarm optimization (PSO) (Abdulwahab, et al., 2024), ant colony optimization (ACO) (Ma, et al., 2021), and genetic algorithms (GA) (Deng, et al., 2023). While they do not guarantee the optimal solution, they provide a favorable balance between accuracy and computational tractability. GA is inspired by natural selection, and evolves a population of candidate solutions via selection, crossover, and mutation (Bohrer and Dorn, 2024; Che, et al., 2025; Deng, et al., 2023). Its strength lies in maintaining solution diversity and reducing the likelihood of local optima. Compared with other heuristic methods, GA offers advantages for feature selection (Li, et al., 2024). It explores multiple solutions in parallel, and its crossover and mutation operators enable systematic recombination of feature subsets, enhancing the discovery of complementary features. These properties make it well suited for high-dimensional, combinatorial feature selection tasks.

However, GA often suffers from slow convergence and reduced effectiveness on high-dimensional datasets. To overcome these challenges, we propose the Helper-Enhanced Feature Selection (HeFS) framework, which augments an initial feature subset with an additional Helper Set of complementary features drawn from the unselected space. Incorporating this helper set enhances classification accuracy, robustness, and feature complementarity, addressing the limitations of existing methods that primarily focus on individually important features. This work makes the following key contributions and evaluates feature selection algorithms across 18 benchmark predictive tasks:

1. We propose Conditional Feature Selection, a general paradigm for augmenting any feature subset with complementary features from the unselected space.

2. We develop three GA optimization strategies for efficient and stable optimization, including biased initialization, ratio-guided mutation, and a robust multi-objective scheme.

3. We demonstrate through 18 benchmark datasets that HeFS consistently improves accuracy and uncovers complementary features missed by baselines.

## 2 Related Work

### 2.1 Feature Selection

Feature selection is a fundamental step in machine learning and data preprocessing, particularly on high-dimensional datasets (Chen, et al., 2020). Its objectives are to identify the most informative features, enhance predictive accuracy, mitigate overfitting, and improve model interpretability. Existing techniques are generally classified into filter, wrapper, and embedded methods (Li, et al., 2017), each offering distinct strengths and limitations, especially when applied to large and complex feature spaces.

Filter methods evaluate feature relevance using statistical measures, information-theoretic scores, or correlation coefficients, independent of any specific learning algorithm. While computationally efficient, early approaches often ignored redundancy among features. Recent advances address this by incorporating redundancy reduction and label correlation modeling. For example, LFFS (Fan, et al., 2022) combines ridge regression, label embeddings, and cosine similarity to suppress redundant features. LCIFS (Fan, et al., 2024) leverages manifold-based regression and adaptive spectral graphs to capture structural label dependencies. Similarly, CCMI

(Zhou, Wang and Zhu, 2022) enhances mutual information by integrating correlation coefficients, improving selection robustness across benchmarks.

Wrapper methods evaluate feature subsets by iteratively training and testing models, employing strategies such as forward selection, backward elimination, and recursive feature elimination (RFE). These methods capture feature dependencies effectively but incur high computational cost. To address scalability, metaheuristic search algorithms, including genetic algorithms (Bohrer and Dorn, 2024), particle swarm optimization (Xue, Zhang and Browne, 2012), and simulated annealing (Shi, et al., 2023), are frequently applied. Examples include CorrACC (Shafiq, et al., 2020), which improves the classification performance of the Internet of Things traffic through a customized evaluation criterion, and GA-based methods for enhanced air pollution prediction (Ul-Saufie, et al., 2022). Hybrid frameworks that combine filter preprocessing with wrapper evaluation, such as FG-HFS (Xu, et al., 2024) and other GA-based hybrids (Bohrer and Dorn, 2024), have demonstrated improved efficiency and robustness, particularly for molecular and multi-label data analysis.

Embedded methods integrate feature selection within the model training process, inheriting the advantages of both filter and wrapper approaches. Representative examples include tree-based models such as Random Forest (Iranzad and Liu, 2024) and regularization-based techniques such as LASSO (Zhang, et al., 2019), which promote sparsity. These methods are generally more computationally efficient than wrappers but may introduce bias towards the characteristics of the underlying learning algorithm.

## 2.2 Genetic Algorithms

Genetic algorithms are population-based optimization methods inspired by the principles of natural selection and genetics (Lambora, Gupta and Chopra, 2019). They iteratively evolve a population of candidate solutions through selection, crossover, and mutation, enabling effective exploration of large, complex search spaces to identify optimal or near-optimal solutions.

GA has proven to be a robust heuristic approach in feature selection, and it can efficiently navigate the combinatorial search space of feature subsets. Feature inclusion is typically encoded as binary strings, where 1 denotes inclusion and 0 denotes exclusion (Bohrer and Dorn, 2024). Early work demonstrated their ability to identify relevant features while discarding irrelevant ones. Notably, (Kohavi and John, 1997) showed that GAs can outperform traditional methods by capturing intricate feature interactions.

Subsequent studies expanded this foundation by integrating application-specific fitness metrics into the GA framework, such as classification accuracy, model complexity, and domain-specific evaluation criteria (Kabir, Shahjahan and Murase, 2011). Multi-objective genetic algorithms (Bohrer and Dorn, 2024; Vijai, 2025) further advanced the field by simultaneously optimizing competing objectives, such as maximizing accuracy while minimizing the number of selected features. Pareto-based selection has been widely adopted to characterize trade-offs between these objectives (Das and Eldho, 2025).

Recent developments have explored hybrid GA strategies, combining GAs with particle swarm optimization or oppositional learning (Che, et al., 2025; Mistry, et al., 2016) to improve convergence and solution quality. Additionally, hybrid multi-objective feature selection (MOFS) methods have been incorporated into ensemble learning frameworks. For example, (Zhou, et al., 2024) proposed a hybrid MOFS approach that generates accurate and diverse classifiers, followed by ensemble selection guided by feature relevance-based diversity metrics, achieving improved balance between accuracy and diversity.

In summary, the role of GAs in feature selection has evolved from simple binary-encoded subset search to sophisticated multi-objective and hybrid frameworks. Ongoing research continues to refine these methods by improving efficiency, scalability, and applicability across domains ranging from molecular data analysis to large-scale, multi-label classification.

**2.3 Multi-objective Optimization**

Multi-objective optimization addresses problems involving two or more conflicting objectives, a common scenario in engineering, finance, and machine learning (Juang and Yeh, 2017; Tian, et al., 2022; Zecchin, et al., 2005). Formally, a general multi-objective problem (Ma, et al., 2023) can be expressed as:

$$\text{Min}_{x \in \Omega} F(x) = [f_1(x), f_2(x), \ldots, f_m(x)], \tag{1}$$

where *x* is a decision vector in the feasible set *Ω*, and $f_i(x)$ represents the $i^{th}$ objective function to be optimized simultaneously.

Multiple strategies exist for solving such problems (Coello, 2006; Sharma and Kumar, 2022), each offering different trade-offs in complexity, interpretability, and flexibility:

1) Weighted Sum Approach: Combines multiple objectives into a single scalar objective by assigning weights to each (Marler and Arora, 2010). While computationally straightforward, its effectiveness depends heavily on the accurate selection of weights to reflect relative importance.

2) Pareto Front Approach: Identifies a set of *non-dominated* solutions (Emmerich and Deutz, 2018), where no solution can be improved in one objective without degrading another. This provides a comprehensive view of trade-offs and is well-suited for decision-making. Representative algorithms include NSGA-II (Vijai, 2025), which uses crowding distance to preserve diversity, and reference-point-based methods (Xia, et al., 2024).

3) ε-Constraint Method: Optimizes one objective while treating the remaining objectives as constraints. This method offers flexibility in exploring trade-offs but typically requires multiple optimization runs to obtain a well-distributed solution set (Sepehri, et al., 2024).

## 3 Method

### 3.1 Problem Formulation and Method Overview

Let $D = \{(x_i, y_i)\}_{i=1}^{n}$ denote a dataset, where $x_i \in \mathbb{R}^d$ is the feature vector of the $i^{th}$ instance, $y_i \in \{0, 1\}$ is its binary label and n denotes the total number of samples. A feature selection algorithm $F$ selects a subset of features $S \subset \{f_1, f_2, \ldots, f_d\}$. A binary classification model $M$ is then trained on $S$, and its performance is evaluated using a metric $\rho(M, D, S)$, which measures predictive quality on dataset $D$.

**Definition 1: Helper Set and Helper Feature**

Given an initial feature subset $S$ selected by a feature selection algorithm $F$ and a performance metric $\rho(M, D, S)$, a helper set $H \subset \{f_1, f_2, \ldots, f_d\} \setminus S$ is defined as a set of complementary features satisfying $\rho(M, D, S \cup H) > \rho(M, D, S)$. Intuitively, features in $H$ may individually have a small or even no association with the class label, but when integrated with $S$, they complement the existing subset and improve the classification performance. When $|F| == 1$, the only feature in $F$ is defined as a helper feature.

**Definition 2: Conditional Feature Selection (CoFS)**

Conditional feature selection task aims to improve the performance of a given feature subset $S$ by discovering helper features from the unselected space.

We define a helper feature as the complementary feature which may have a small association with the class label but can improve the classification performance of a feature subset with its integration. Let $S$ be the initial feature subset selected by the algorithm $F$. We aim to identify an additional set of features $S' \subset \{f_1, f_2, \ldots, f_d\} \setminus S$ such that $|S' \cup S| \leq d$, and the combined feature set $S' \cup S$ yields a higher $\rho$ for $M$ on $D$. This defines a conditional feature selection problem, where the task is to improve the performance of a given feature subset $S$ by discovering *complementary* features from the unselected space.

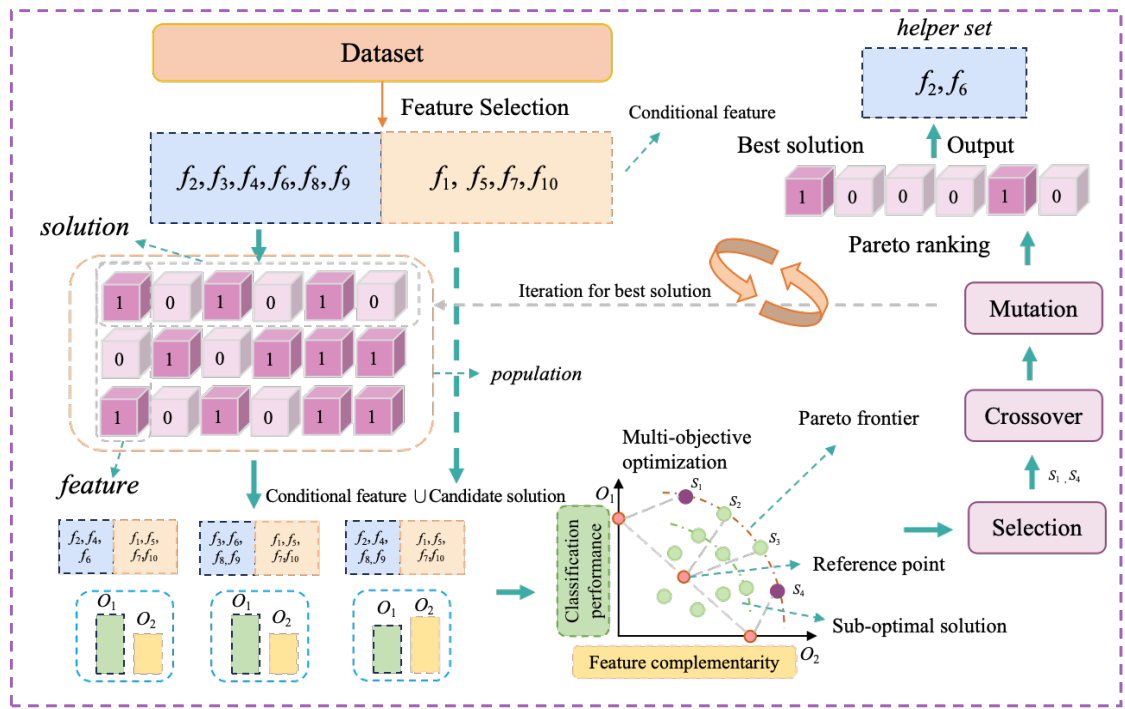

Figure 1. The overall workflow of the Helper-Enhanced Feature Selection (HeFS) framework.

We implement our algorithm Helper-Enhanced Feature Selection (HeFS) in this setting:

- $S$ is the subset initially selected by an existing feature selection method and is termed the conditional set.

- $S'$ is the complementary features identified by HeFS to augment $S$ and is termed the helper set.

Our approach employs a multi-objective genetic algorithm with a reference-point mechanism and niching strategy to simultaneously optimize:

- Predictive performance by maximizing the classification accuracy (or relevant metric) of $S \cup S'$.

- Feature complementarity by maximizing the degree to which $S'$ contributes non-redundant, informative features relative to $S$.

The conditional set $S$ acts as a guidance set during the search and it steers exploration toward promising regions of the feature space. Customized genetic operators are designed to iteratively generate and refine candidate subsets over multiple generations. The algorithm produces a set of non-dominated solutions forming a Pareto front, each representing a trade-off between accuracy and complementarity. The subset with the highest prediction accuracy is selected as the final output.

Figure 1 presents a high-level overview of the HeFS framework. Subsequent subsections describe the customized genetic operators, dimensionality reduction strategies, and the conditional feature selection process that enables the discovery of additional informative features through multi-objective optimization.

## 3.2 The HeFS Framework

The Helper-Enhanced Feature Selection (HeFS) framework is designed to enhance an existing feature subset by identifying additional complementary features (referred to as the *Helper Set*) from the unselected space. This design is motivated by the observation that conventional feature selection methods may overlook weakly relevant but complementary features, which may further improve classification performance when integrated with an existing subset.

Firstly, we need to define the best helper set among a population of feature subsets under the conditional set *S*.

---
**Algorithm 1: BestHelperSet**

---
**Input**: Dataset *D*, target *Y*, conditional set *S*, and a population of feature subsets *P'*.
**Output**: Helper set *S'*, final performance *R*
1:   *S'* = *P'*[0]; *R* = **EvaluateModel**(*S'*∪*S*);
2:   **for** *i* = 0 to (|*P'*|-1) **do**
3:     *TempR* = **EvaluateModel**(*P'*[*i*]∪*S*)
4:     **if** *TempR* > *R* **do**
5:       *S'* = *P'*[*i*]; *R* = *TempR*;
6:     **end if**
7:   **end for**
8:   **return** *S'*, *R*

---

An overview of the workflow is shown in **Figure 1**, and the corresponding pseudo-code is provided in **Algorithm 2**.

---
**Algorithm 2: Helper-Enhanced Feature Selection (HeFS)**

---
**Input**: Dataset *D*, Target *Y*, Feature selection algorithm *F*,
**Output**: Helper set *S'*, final performance *R*
9:   *S* = *F*(*D*, *Y*); $S_1$ = *D*\\*S*; *t* = 0;
10:  $P_h(0)$ = **SelectiveActivation**($S_1$) # Initialize the population of helper sets
11:  *P*(0) = **ParetoSolutions**(*S*, $P_h(0)$) # Extract the helper sets on the Pareto front
12:  **while** *t* < *T* **do**
13:    $P_s$ = **SelectionOperator**($P(t)$)
14:    $P_c$ = **CrossOperation**($P_s$)
15:    $P_m$ = **IntelligentMutation**($P_c$)
16:    *P*(*t*+1) = **ParetoSolutions**(*S*, $P_m$ ∪*P*(0)) # Keep the previous best solutions
17:    *t* = *t*+1
18:  **end while**
19:  *S'*, R = **BestHelperSet**(*D*, *Y*, *S*, *P*(*t*))
20:  **return** *S'*, *R*

---

Given a dataset *D*, target labels *Y*, and a baseline feature selection algorithm *F*, the proposed algorithm HeFS begins by obtaining an initial conditional set *S* using *F*. A selective activation initialization strategy SelectiveActivation() is used to construct the initial population $P_h(0)$. SelectiveActivation() initializes the population by selectively activating informative and sparse feature subsets, with details provided in Section 3.3.

The optimization process is driven by a multi-objective genetic algorithm guided by Pareto dominance (Imani, et al., 2024). At each iteration, a candidate helper set $H$ is evaluated in combination with S, i.e., EvaluateModel($S \cup H$). EvaluateModel() is an evaluation function that returns the predictive performance of the model built on the given feature subset. The population is evolved through three genetic operators, i.e., SelectionOperator(), CrossOperation(), and IntelligentMutation(). IntelligentMutation() adaptively mutates individuals in a ratio-guided manner to maintain sparsity and diversity, with details provided in Section 3.3. At the end of each iteration, only the helper sets on the Pareto front are kept in the population using the function ParetoSolutions().

The process terminates after the pre-set $T$ iterations. HeFS returns the helper set $S'$ from the final population $P(t)$ that achieves the highest classification accuracy in combination with $S$. The above-mentioned functions are described in the following sections.

The SelectionOperator() selects a subset of individuals from the population based on Pareto dominance and, optionally, diversity metrics. First, the population is divided into Pareto fronts, with individuals in higher fronts being prioritized. Within each front, optional diversity metrics, such as reference point-based sorting, are applied to maintain a diverse solution set. Individuals are then selected from the top fronts until the desired population size $k$ is achieved.

The CrossOperation() performs genetic recombination between two parent individuals to generate offspring. In our implementation, we employ single-point crossover: a random crossover point is chosen along the feature vector, and the segments of the two parents are exchanged to produce two new offspring. Formally, given two parents, parent1 and parent2, CrossOperation(parent1, parent2) returns two offspring by concatenating the first part of one parent with the second part of the other.

## 3.3 Selective Activation Initialization Strategy

To mitigate slow convergence and unrepresentative initial populations in high-dimensional feature spaces, the HeFS framework incorporates a Selective Activation Initialization Strategy, i.e., the function SelectiveActivation() in the Algorithm 2. This strategy combines two steps: Sample Clustering and Biased Sampling, designed to ensure both diversity and sparsity in the initial population.

The Sample Clustering step aims to improve representativeness by clustering the samples based on cosine distance. Two samples are assigned to the same cluster $C_k$ if their cosine distance is below a predefined threshold $\delta = 0.1$. To reduce redundancy, only one representative sample is retained per cluster, randomly selected as $x_k$. This yields a reduced and diverse dataset $D'$, which serves as the foundation for population initialization:

$$dist(x_i, x_j) = 1 - \frac{x_i \cdot x_j}{\| x_i \| \cdot \| x_j \|} \tag{1}$$

$$dist(x_i, x_j) < \delta \tag{2}$$

$$x_k^* \sim Uniform(C_k) \tag{3}$$

$$D' = \{x_1^*, x_2^*, \dots x_m^*\}, m < n \tag{4}$$

This clustering and deduplication process ensures that initialization captures diverse sample information without redundancy for the improved efficiency in subsequent optimization.

Following clustering, the Biased Sampling step controls the number of active features in the initial population. This encourages sparse but informative subsets that serve as better starting points for the evolutionary process. The mechanism is parameterized by a minimum ratio $R_{min}$, a maximum ratio $R_{max}$, and a scaling factor $Scaler$. A value $R \in [0, 1]$ is drawn from a uniform distribution, and the activation ratio is computed as:

$$s = R_{min} + (R_{max} - R_{min}) \cdot e^{-Scaler} \tag{5}$$

$$ratio = \min(R_{max}, \max(R_{min}, s)) \qquad (6)$$

This formulation biases the ratio toward smaller values, generating subsets with reduced dimensionality. By starting with sparse solutions, the search process is accelerated and guided toward tractable regions of the feature space. Given a feature space of size $n$, each candidate solution is encoded as a binary activation vector of length $n$, where the number of activated features is determined by $\lfloor n \times ratio \rfloor$. The activated positions are initially assigned as 1 while the rest remain 0, followed by a stochastic permutation to eliminate positional bias and enhance population diversity. By iterating this procedure for the specified population size, a diverse yet systematically constrained set of candidate feature subsets is generated, and serves as the initialization pool for subsequent evolutionary search. This formulation biases the ratio toward smaller values, and generates subsets with reduced dimensionality. By starting with sparse solutions, the search process is accelerated and guided toward tractable regions of the feature space.

Notably, the same Biased Sampling mechanism is reused in the Intelligent Mutation operator of HeFS, ensuring consistent control over feature activation throughout optimization. By combining sample-level diversity with feature-level sparsity, the Selective Activation Initialization Strategy provides a principled and efficient basis for initializing populations in the HeFS framework.

### 3.4 Ratio-Guided Mutation Strategy

We further introduce a ratio-guided mutation strategy (function IntelligentMutation()), regulated by a target ratio threshold and the biased sampling algorithm. In conventional genetic algorithms, mutation is typically performed by randomly selecting one or more positions to flip. However, excessive randomness can disrupt the distribution of selected features, and hinder stable convergence toward high-quality solutions.

To address this, we define a target ratio using the biased sampling strategy (see Equations (5)–(6)), which guides mutation to preserve population balance while maintaining diversity. The current ratio of an individual is defined as the proportion of selected features (i.e., those encoded as 1) relative to the total number of features.

Let the current ratio and the target ratio be $r_{current}$ and $r_{target}$.

$$r_{current} = \frac{1}{d}\sum_{i=1}^{d} I[x_i = 1] \qquad (7)$$

Given $r_{target}$ from (5)-(6),

$$\Delta r = |r_{current} - r_{target}| \qquad (8)$$

The indicator function $I[z = 1]$ is 1 if z equals to 1, otherwise 0.

Case 1: (near/above target): if $\Delta r < \varepsilon$ or $r_{current} > r_{target}$, one selected (1) and one non-selected (0) features are randomly chosen for inversions.

Case 2 (below target): if $r_{current} \leq r_{target} - \varepsilon$, $for\ i = 1, \dots, d$, define the adjustment probability

$$P_{adjust} = \min\{1,\ \max\{0,\ r_{target} - r_{current}\}\} \qquad (9)$$

and apply a probabilistic bit flip independently to each feature (position), $P_{adjust}$ is updated in real time, and $P_{adjust}^{(j)}$ represents the adjustment probability when the current individual is updated to the $i^{th}$ position:

$$x_i' = \begin{cases} 1 - x_i & with\ probability\ P_{adjust}^{(j)} \\ x_i & with\ probability\ 1 - P_{adjust}^{(j)} \end{cases} \qquad (10)$$

This strategy ensures that the number of selected features evolves smoothly toward the target ratio while preserving exploration capacity across generations.

### 3.5 Approach for Multi-objective Optimization

In the HeFS framework, the evaluation of candidate helper feature subsets is formulated as a multi-objective optimization problem. We design a dual-objective fitness function that jointly considers classification accuracy and feature complementarity, forming the basis for optimization through reference-point-guided and niching-based evolutionary strategies.

The overall fitness is defined as $Fitness = (Fitness_1, Fitness_2)$. The first objective component evaluates the predictive utility of the combined feature set $S \cup H$, where $S$ is the conditional feature subset and $H$ is the candidate helper set. A classifier is trained on $S \cup H$, and its validation accuracy is taken as the first fitness score: $Fitness_1 = Accuracy(S \cup H)$. The second objective component measures the complementarity between $H$ and $S$. We compute mutual information (MI) between features in $H$ and those in $S$, and define the complementarity score as: $Fitness_2 = 1 - \frac{\text{mean}(MI(S \cup H))}{max(MI(S \cup H))}$.

Conventional multi-objective genetic algorithms often rely on crowding distance to preserve solution diversity along the Pareto front (Sağlican and Afacan, 2023). However, distance-based measures lose effectiveness due to the concentration of pairwise distances in high-dimensional spaces (Kumari and Jayaram, 2017). This effect can be explained by the Central Limit Theorem (CLT) (Angiulli, 2018): for a sequence of i.i.d. random variables $<X_1, X_2, \ldots, X_n>$ with mean $\mu$ and variance $\sigma^2$, the standardized sum $Z_n = \frac{S_n - n\mu}{\sqrt{n\sigma^2}}$, where $S_n = \sum_{i=1}^{n} X_i$ converges in distribution to a standard normal distribution as $n \to \infty$. This is expressed as:

$$\lim_{n \to \infty} P(Z_n \leq z) = \lim_{n \to \infty} P\left(\frac{S_n - n\mu}{\sqrt{n\sigma^2}} \leq z\right) = \Phi(z) \qquad (11)$$

where $\Phi(z)$ denotes the standard normal cumulative distribution function. Interpreting each objective dimension as a random variable in the context of multi-objective optimization, the pairwise distances between solutions can be approximated as sums of i.i.d. variables. By CLT, these distances concentrate around their mean, which reduce their discriminative power in higher dimensions.

Geometrically, the expected Euclidean distance between two random points in a $d$-dimensional unit hypercube is $\mathbb{E}[\|X - Y\|] = \sqrt{d/6}$, and its variance $Var[D] = O(1/d)$ decreases as $d$ increases (François, Wertz and Verleysen, 2007), which further confirms the distance concentration phenomenon. Consequently, crowding distance becomes ineffective in distinguishing sparse from dense regions along the Pareto front.

To overcome these limitations, we adopt a reference-point-based selection with niching strategy. All Pareto-optimal solutions are normalized to a common scale, and a set of uniformly distributed reference points is generated in the objective space. Each solution is assigned to its nearest reference point using Euclidean distance. Niche counts are then computed per reference point, and solutions associated with less crowded niches are preferentially selected. This promotes diversity and mitigates premature convergence. As illustrated in Figure 1, solutions $s_1$ and $s_2$ are retained due to their association with sparsely populated niches, while in a more crowded niche, whether a solution is retained depends on the desired number of selections: if more solutions are needed, one solution (e.g., $s_3$) is randomly chosen from the more crowded niche; otherwise, the rest are discarded.

To further ensure balanced solution distribution, we propose an adaptive partitioning strategy. Instead of using a fixed partition scheme, the number of partitions is dynamically determined by the size of the current Pareto front:

$$P = \max\left(1, \left\lfloor \log(|F| + 1) \times \sqrt{|F|} \right\rfloor\right) \tag{12}$$

where $|F|$ is the number of solutions on the front. The logarithmic term ensures finer granularity when few solutions are present, while the square-root term moderates growth for larger fronts and prevents over-fragmentation. This hybrid design balances resolution and stability, yielding more accurate ranking and improved adaptability.

Together, the reference-point mechanism, niching strategy, and adaptive partitioning form a robust and scalable screening for the candidate helper set solutions on the Pareto front (i.e., function ParetoSolutions()) for multi-objective optimization within the HeFS framework, particularly under high-dimensional and complex feature selection scenarios.

## 4. Experiment Settings

### 4.1 Datasets

We evaluate the proposed method on a diverse collection of datasets drawn from the UCI Machine Learning Repository (Kelly, Longjohn and Nottingham, 2025), the DBC Repository (https://leo.ugr.es/elvira/DBCRepository/) (Cano, Masegosa and Moral, 2025), the Scikit Feature Datasets (https://jundongl.github.io/scikit-feature/datasets.html) (Li, et al., 2017), the FS-DB database (https://github.com/lyceia/FS-DB) (Wang, Luo and Yao, 2024), and NCBI Repository (Barrett, et al., 2012) (Table 1). In total, 20 datasets spanning different domains were used. The datasets vary substantially in scale: the number of features ranges from 36 to over 50,000, while the number of samples ranges from fewer than 100 to several thousand.

To ensure reliable evaluation and reduce the effect of random partitioning on generalization, all experiments were conducted using 5-fold cross-validation.

**Table 1. Summarizations of the datasets.** The columns "ID", "Dataset", "Src", and "Domain" give the abbreviated ID, full name, source, and domain of each dataset. The columns "Classes" gives the number of classes, "Features" denotes the dimensionality and "Samples" denotes the number of samples.

| Dataset | Src | Domain | Classes | Samples | Features |
|---|---|---|---|---|---|
| PenglungEW | UCI | Medical | 7 | 73 | 325 |
| Satellite | UCI | Climate and Environment | 6 | 4435 | 36 |
| Semeion | UCI | Computer Science | 2 | 1593 | 265 |
| Spambase | UCI | Computer Science | 2 | 4601 | 57 |
| WaveformEW | UCI | Physics and Chemistry | 3 | 5000 | 40 |
| Leukemia1 | FS-DB | Biological Data | 3 | 72 | 5327 |
| Leukemia2 | FS-DB | Biological Data | 3 | 72 | 11225 |
| DLBCL | DBC | Biological Data | 2 | 77 | 5469 |
| Prostate_Tumor | FS-DB | Biological Data | 2 | 102 | 10509 |
| Prostate1 | FS-DB | Biological Data | 2 | 102 | 5966 |
| MLL | FS-DB | Biological Data | 3 | 72 | 12582 |
| GLI-85 | Scikit | Biological Data | 2 | 85 | 22283 |
| Lung | DBC | Biological Data | 5 | 203 | 12600 |
| LungCancer | DBC | Biological Data | 2 | 181 | 12533 |

| Ovarian | DBC | Biological Data | 2 | 253 | 15154 |
| Prostate-GE | Scikit | Biological Data | 2 | 102 | 5966 |
| GSE64951 | NCBI | Biological Data | 2 | 94 | 54675 |
| Toxicity | UCI | Molecule | 2 | 171 | 1203 |

## 4.2 Comparison Algorithms and Experimental Setup

To ensure robust evaluation, each dataset was tested over 10 independent random runs and 5-fold cross-validation strategy. Classification performance was assessed through the EvaluateModel() function, which in our experiments was instantiated as a k-nearest neighbors classifier ($k = 5$). The performance metric was defined as the average classification accuracy across the 5 folds. The population size of the GA was set to 30 and the maximum number of iterations to 100. Parameters in the Biased Sampling mechanism ($R_{min}$, $R_{max}$, $Scaler$) were tuned empirically as detailed in Section 5.1.

Evaluation was conducted using multiple metrics, including AUC, Precision (Prec), Recall (Rec), and Accuracy (Acc), to provide a comprehensive assessment of predictive performance.

For comparative analysis, we benchmarked our method against a broad spectrum of feature selection approaches, including

- **Classical baselines:** Random Forest (RF), Decision Tree (DT), Lasso, Mutual Information (MutualInfo), and Ridge Regression (Ridge).

- **Recent state-of-the-art methods:** HGSA (Taradeh, et al., 2019), SBOA (Arora and Anand, 2019), VCOA (de Souza, et al., 2020), FTGGA (Deng, et al., 2023), MGWO (Pan, Chen and Xiong, 2023), and BHOA (Pashaei, Pashaei and Mirjalili, 2025).

# 5. Results and Discussion

## 5.1 Hyperparameter Tuning and Sensitivity Analysis

| | | | $R_{max}$ | | | | | | | $R_{max}$ | | | |
|---|---|---|---|---|---|---|---|---|---|---|---|---|---|
| Acc | | | 0.01 | 0.03 | 0.05 | 0.07 | Acc | | | 0.01 | 0.03 | 0.05 | 0.07 |
| MLL | $R_{min}$ | 0.2 | 0.9266 | 0.9264 | 0.9319 | 0.9304 | Leukemia2 | $R_{min}$ | 0.2 | 0.9263 | 0.9279 | 0.9198 | 0.9293 |
| | | 0.3 | 0.9278 | 0.9308 | **0.9346** | 0.9264 | | | 0.3 | 0.9221 | 0.9180 | **0.9304** | 0.9224 |
| | | 0.4 | 0.9279 | 0.9317 | 0.9306 | 0.9306 | | | 0.4 | 0.9223 | 0.9250 | 0.9262 | 0.9189 |
| | | 0.5 | 0.9277 | 0.9319 | 0.9289 | 0.9294 | | | 0.5 | 0.9181 | 0.9193 | 0.9205 | 0.9247 |
| PenglungEW | $R_{min}$ | 0.2 | 0.9277 | 0.9314 | 0.9385 | 0.9285 | Prostate-GE | $R_{min}$ | 0.2 | 0.9236 | 0.9245 | 0.9315 | 0.9285 |
| | | 0.3 | 0.9210 | 0.9290 | **0.9428** | 0.9326 | | | 0.3 | 0.9216 | 0.9275 | **0.9324** | 0.9305 |
| | | 0.4 | 0.9192 | 0.9256 | 0.9272 | 0.9261 | | | 0.4 | 0.9234 | 0.9265 | 0.9249 | 0.9284 |
| | | 0.5 | 0.9222 | 0.9275 | 0.9271 | 0.9319 | | | 0.5 | 0.9217 | 0.9285 | 0.9275 | 0.9236 |

(a)

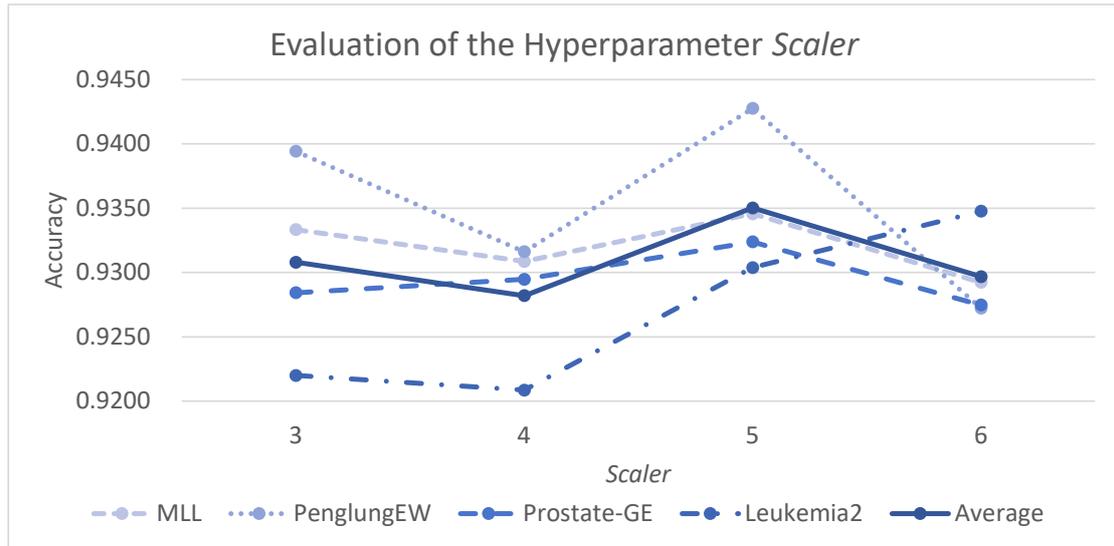

(b)

**Figure 2. Tuning of the Hyperparameters of the HeFS framework.** (a) The classification accuracies of tuning the hyperparameter $R_{min}$ and $R_{max}$ over the four representative datasets, i.e., MLL, PenglungEW, Leukemia2, and Prostate-GE. (b) The classification accuracies of tuning the hyperparameter $Scaler$ over the four representative datasets, i.e., MLL, PenglungEW, Leukemia2, and Prostate-GE.

We examined the impact of key hyperparameters in the Biased Sampling mechanism through a sensitivity analysis of three parameters: $R_{min}$, $R_{max}$, and $Scaler$. These parameters regulate the range and probability of feature selection in the evolutionary search. Experiments were performed on four representative datasets (MLL, PenglungEW, Leukemia2, and Prostate-GE) with 10 independent random runs. The mean classification accuracy was used as the performance metric.

We first jointly tuned $R_{min} \in \{0.01, 0.03, 0.05, 0.07\}$ and $R_{max} \in \{0.2, 0.3, 0.4, 0.5\}$. Figure 2 (a) reports the overall classification accuracies across four representative datasets. The configuration $(R_{min} = 0.05, R_{max} = 0.3)$ consistently achieved the strongest performance, with accuracies of 0.9346 on MLL, 0.9428 on PenglungEW, and 0.9324 on Prostate-GE, as well as competitive performance on Leukemia2 (0.9304). Averaged across all datasets, this setting reached a mean accuracy of 0.9350, outperforming other combinations of the two hyperparameters. These results suggest that setting $R_{min}$ to overly small values (e.g., 0.01) leads to insufficient feature diversity, whereas excessively large values (e.g., 0.5) introduce excessive randomness that undermines the optimization process. The selected configuration $(R_{min} = 0.05, R_{max} = 0.3)$ achieves a balanced trade-off between exploration and exploitation.

With $R_{min} = 0.05$ and $R_{max} = 0.3$ fixed, we tuned the $Scaler$ parameter, which controls the degree of exponential bias toward smaller sampling ratios, and thereby influences the sparsity of selected features. Candidate values $\{3, 4, 5, 6\}$ were evaluated for the hyperparameter $Scaler$ over the four representative datasets, as shown in Figure 2 (b). The best overall performance was obtained with $Scaler = 5$, yielding the highest or near-highest accuracies on MLL (0.9346), PenglungEW (0.9428), and Prostate-GE (0.9324). Although $Scaler = 6$ achieved the best score on Leukemia2 (0.9348), it was less stable across the other datasets.

These findings suggest that $(R_{min} = 0.05, R_{max} = 0.3, Scaler = 5)$ offers the most effective and consistent configuration. This setting induces a moderate bias toward sparsity, and enhances the discovery of informative features while avoiding premature convergence. The stability of performance across datasets demonstrates the generalizability and robustness of these hyperparameter choices.

## 5.2 Ablation Experiment

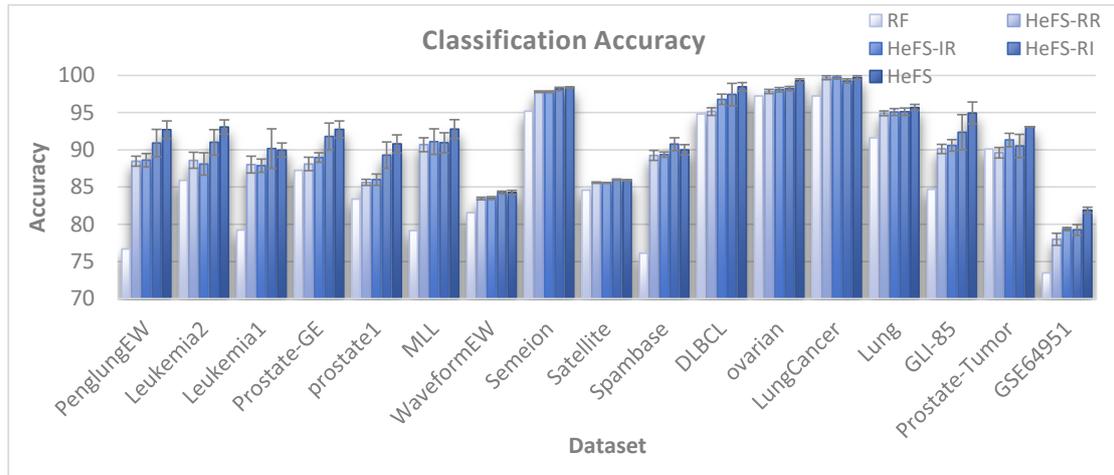

(a)

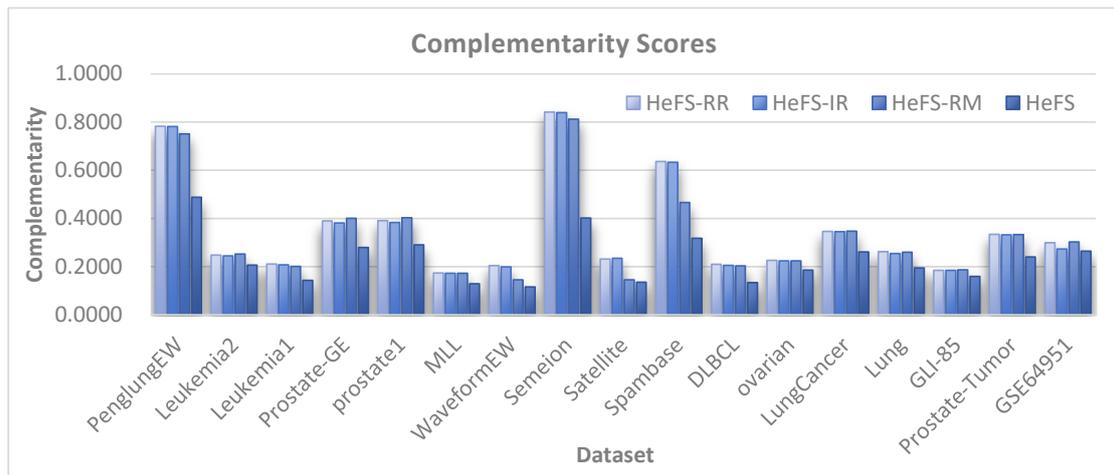

(b)

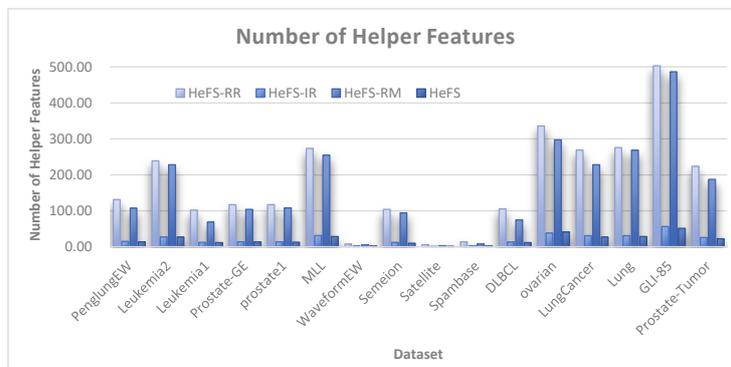

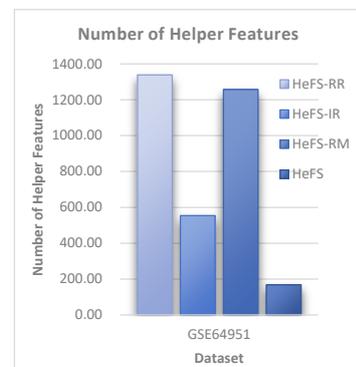

(c)                                                           (d)

Figure 3. Ablation Experiments. (a) The classification accuracies of the four variants of HeFS are evaluated over 10 runs and the column heights show the averaged accuracies. The error bars illustrate the standard deviations. The classifier RF serves as the baseline. (b) The averaged complementarity scores of the helper sets detected by the four variants of HeFS. (c) Average number of helper features detected by the four HeFS variants across the first 16 benchmark datasets, and (d) results on the gastric cancer dataset (GSE64951). The histograms are shown separately because the number of helper features in the gastric cancer dataset (GSE64951) is substantially larger than in the other datasets.

To quantify the contributions of the biased initialization and ratio-guided mutation in HeFS, we evaluate four variants: 1) HeFS-RR: random initialization + random mutation (baseline). 2) HeFS-IR: Selective Activation Initialization + random mutation (isolates initialization). 3) HeFS-RM: random initialization + ratio-guided mutation (isolates mutation). 4) HeFS: full method with both components.

We report mean accuracy over 10 independent runs for each dataset (Figure 3 (a)). HeFS attains the highest average accuracy on 13/17 datasets, with significant gains on high-dimensional biomedical data (e.g., DLBCL 98.47%, Ovarian 99.29%, GLI-85 94.94%). On Leukemia1, Satellite, Spambase, and LungCancer, HeFS remains close to the best variant. These results indicate that combining both components yields consistent improvements across domains.

Complementarity of a helper set is denoted as the metric $Fitness_2$, and a smaller complementarity score suggests that this helper set delivers a better contribution to the core feature set. Figure 3 (b) reports average complementarity scores of the four variants of HeFS. The full method HeFS achieves the lowest complementarity scores on all datasets, with substantial reductions on PenglungEW (0.4883), Semeion (0.4017), Spambase (0.3178), and GLI-85 (0.1593), evidencing more diverse and informative subsets.

Figure 3 (c) evaluates the numbers of helper features detected by the four variants of HeFS. HeFS selects markedly fewer helper features while maintaining superior accuracy, e.g., PenglungEW (13.9 vs. 131.0 for HeFS-RR) and Prostate-GE (13.0 vs. 116.7). A smaller number of selected helper features improves interpretability and reduces computational cost.

Supplementary Figure S1 shows accuracy versus iterations. The green dashed line is the RF baseline using a fixed set of 20 features. All variants surpass the baseline on most numbers of helper features. HeFS converges faster and to higher accuracy, underscoring the benefit of coupling initialization with ratio-guided mutation.

## 5.3 Comparison with Feature Selection Algorithms

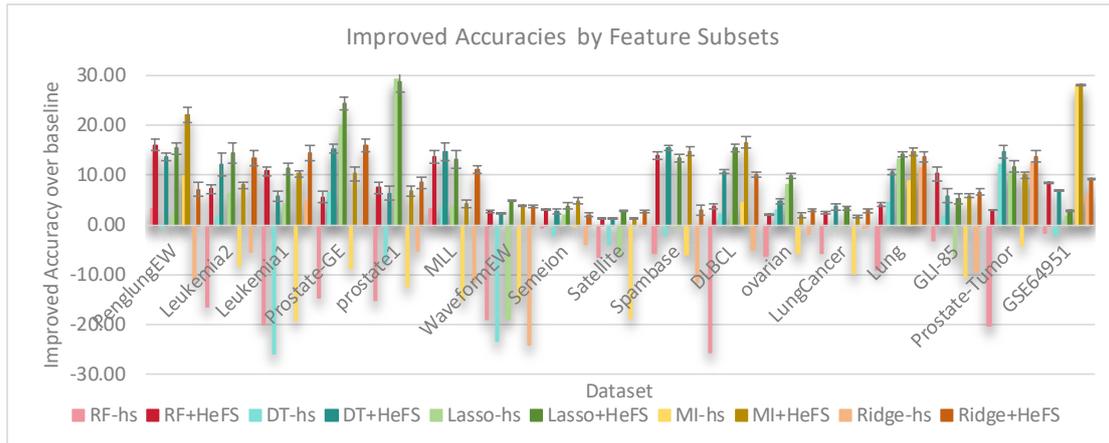

(a)

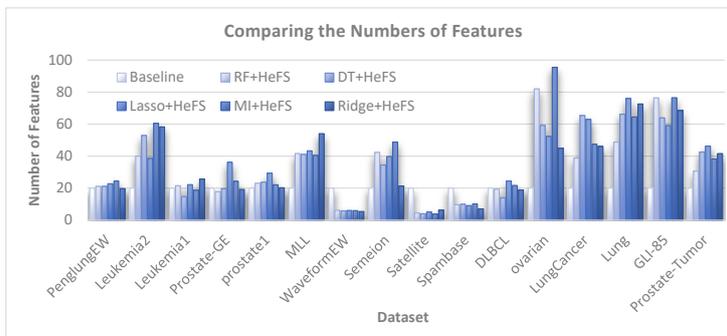

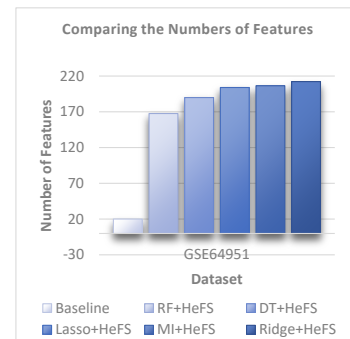

(b)                                        (c)

Figure 4. Performance Comparison with the Baseline Feature Selection Algorithms. (a) Improvements of classification accuracies (%) over baseline algorithms, i.e., RF, DT, Lasso, MI, and Ridge. The helper-only variants are denoted by the suffix "-hs", and the combined methods have the suffix "+HeFS" across 16 datasets (mean ± standard deviation). (b) The numbers of features selected by the feature selection algorithms are evaluated across the first 16 benchmark datasets, and (c) results on the gastric cancer dataset (GSE64951). The baseline feature selection algorithms choose 20

features, while the helper feature sets selected by HeFS are denoted by the suffix "+HeFS". The histograms are shown separately because the number of helper features in the gastric cancer dataset (GSE64951) is substantially larger than in the other datasets.

To further validate the effectiveness of *HeFS*, we compared its performance against five widely used feature selection algorithms, i.e., Random Forest (RF), Decision Tree (DT), Lasso, Mutual Information (MI), and Ridge Regression. For each baseline, three strategies were evaluated: the baseline method alone, the helper set alone, and the *combined method*, where features selected by the baseline were augmented with helper features identified by *HeFS*. Classification accuracies of the combined methods were measured by mean ± standard deviation over 10 independent runs across 16 datasets. Figure 4 (a) shows that the baseline feature selection algorithms, when used in isolation, often underperform due to their inability to capture complementary information. The helper features alone yield modest accuracy, confirming that while they are not strong predictors individually, they contain complementary signals. When integrated with baseline-selected features in the HeFS framework, performance improves substantially and consistently across all datasets.

HeFS achieves the improved accuracies across all the baseline feature selection algorithms on all 17 datasets in Figure 4 (a), with particularly notable gains on challenging high-dimensional benchmarks such as PenglungEW, Leukemia2, Leukemia1, Prostate-GE, prostate1, and GLI-85, where improvements over all the baselines exceed 5 percentage points. These results demonstrate that HeFS effectively exploits feature complementarity to enhance predictive power while maintaining robustness across diverse domains.

Figure 4 (b) reports the number of selected features. The baseline feature selection algorithms select a fixed number of 20 features, while HeFS dynamically adapts the helper subset size by combining baseline and helper-selected features. Across datasets, HeFS often selects fewer but more informative features, with subsets typically ranging between 11 and 30. On datasets such as GLI-85 and Ovarian, more features are retained when necessary to preserve complementarily informative signals. Conversely, on WaveformEW and Satellite, HeFS selects very compact subsets (as few as 4–6 features) without loss of accuracy. These results confirm that HeFS adapts feature subset size to dataset complexity, balancing compactness with predictive strength.

We further evaluated accuracy convergence over 100 iterations (Supplementary Figure S2). Across most datasets, HeFS rapidly surpasses the baseline algorithms and continues to improve, demonstrating its capacity to refine feature subsets iteratively. Even when initial accuracy is comparable to or slightly below the baselines, HeFS consistently uncovers complementary helper features, resulting in superior final performance.

Supplementary Figure S3 presents box plots of classification improvements brought by the helper set over the five baselines. Accuracy gains are consistently positive, with most median values significantly above zero. The narrow interquartile ranges demonstrate the stability of the improvements across independent runs. These results confirm that the helper features provide complementary value and that HeFS integrates them effectively to achieve reliable performance gains.

### 5.4 Comparison with State-of-the-Art Methods

Table 2. Comparison with the state-of-the-art (SOTA) feature selection algorithms. Classification accuracy (%) of SOTA methods, and combined HeFS-augmented methods across 16 datasets (averaged accuracy).

| Dataset | HGSA | HGSA+HeFS | SBOA | SBOA+HeFS | VCOA | VCOA+HeFS |
|---|---|---|---|---|---|---|
| PenglungEW | 82.10 | 83.97 | 83.52 | 87.85 | 75.33 | 89.75 |
| Leukemia2 | 70.57 | 83.19 | 72.29 | 84.77 | 73.62 | 91.05 |
| Leukemia1 | 79.24 | 87.82 | 71.90 | 81.61 | 69.33 | 90.13 |
| prostate1 | 75.57 | 82.70 | 77.62 | 85.25 | 91.24 | 92.87 |
| Prostate-GE | 79.52 | 87.13 | 87.24 | 93.10 | 83.33 | 91.31 |
| MLL | 87.33 | 93.21 | 75.05 | 87.31 | 83.52 | 94.89 |
| WaveformEW | 78.90 | 83.52 | 75.18 | 82.16 | 72.56 | 82.08 |
| Semeion | 96.30 | 96.29 | 96.17 | 97.00 | 94.79 | 98.92 |
| Satellite | 83.09 | 85.13 | 82.44 | 85.77 | 82.62 | 86.13 |
| Spambase | 73.98 | 86.84 | 75.79 | 88.70 | 74.68 | 90.46 |
| DLBCL | 85.67 | 89.37 | 90.75 | 93.73 | 88.33 | 96.57 |
| ovarian | 95.67 | 97.00 | 95.27 | 97.79 | 95.65 | 99.05 |
| LungCancer | 97.78 | 99.00 | 93.36 | 99.50 | 93.92 | 99.28 |
| Lung | 88.70 | 94.07 | 88.67 | 94.72 | 85.24 | 95.01 |
| GLI-85 | 82.35 | 89.88 | 82.35 | 89.29 | 81.18 | 92.35 |
| Prostate-Tumor | 78.38 | 84.86 | 81.43 | 86.71 | 78.48 | 90.66 |
| GSE64951 | 71.23 | 79.91 | 62.81 | 70.23 | 59.47 | 71.57 |
| Dataset | FTGGA | FTGGA+HeFS | MGWO | MGWO+HeFS | BHOA | BHOA+HeFS |
| PenglungEW | 86.29 | 89.45 | 86.38 | 93.20 | 87.62 | 94.00 |
| Leukemia2 | 69.24 | 81.98 | 77.43 | 91.92 | 81.71 | 95.47 |
| Leukemia1 | 69.24 | 83.74 | 84.57 | 90.36 | 87.52 | 91.71 |
| prostate1 | 82.57 | 87.77 | 86.38 | 92.73 | 90.24 | 93.10 |
| Prostate-GE | 74.57 | 85.22 | 83.38 | 91.99 | 87.24 | 93.46 |
| MLL | 86.10 | 94.41 | 97.14 | 97.25 | 91.43 | 94.62 |
| WaveformEW | 83.86 | 84.63 | 56.32 | 81.03 | 80.74 | 83.85 |
| Semeion | 97.80 | 96.62 | 95.42 | 98.85 | 100.00 | 100.00 |
| Satellite | 84.71 | 85.13 | 83.22 | 85.26 | 85.66 | 86.64 |
| Spambase | 77.11 | 89.90 | 57.64 | 89.44 | 86.89 | 91.35 |
| DLBCL | 87.00 | 94.39 | 87.00 | 100.00 | 97.50 | 99.88 |
| ovarian | 96.05 | 97.87 | 98.82 | 99.33 | 99.21 | 99.33 |
| LungCancer | 98.33 | 98.89 | 97.79 | 99.56 | 97.79 | 99.72 |
| Lung | 91.10 | 94.27 | 89.65 | 95.40 | 92.10 | 96.13 |
| GLI-85 | 84.71 | 90.24 | 94.12 | 96.00 | 90.59 | 94.35 |
| Prostate-Tumor | 81.38 | 85.28 | 86.29 | 91.65 | 86.10 | 92.80 |
| GSE64951 | 68.01 | 75.16 | 77.72 | 80.69 | 78.83 | 82.83 |

To assess the competitiveness of HeFS, we compared it with six state-of-the-art GA-based feature selection algorithms: HGSA (Taradeh, et al., 2019), SBOA (Arora and Anand, 2019), VCOA (de Souza, et al., 2020), FTGGA (Deng, et al., 2023), MGWO (Pan, Chen and Xiong, 2023), and BHOA (Pashaei, Pashaei and Mirjalili, 2025). Table 2 reports classification accuracy across 17 benchmark datasets. In nearly all cases, the HeFS-augmented variants outperform their baselines, confirming that helper-selected features consistently contribute complementary information. For example, on the DLBCL dataset, BHOA alone achieves 97.50% accuracy, while the helper features alone reach only 73.18%. However, the combined BHOA+HeFS attains 99.88%, demonstrating a substantial performance boost. A single exception occurs on the Semeion dataset, where FTGGA achieves 97.80% while FTGGA+HeFS slightly decreases to 96.62%. This minor degradation likely arises from redundancy or overfitting due to feature expansion in a dataset with limited tolerance to noise. Overall, the improvements delivered by HeFS are consistent, substantial, and transferable across diverse datasets, highlighting the broad applicability of the framework.

On the gastric cancer dataset GSE64951, applying the helper-selection strategy (-hs) leads to noticeable improvements in most algorithms, confirming that the features identified by HeFS effectively complement subsets generated by other methods. This demonstrates that HeFS not only recovers overlooked yet informative features but also integrates robustly with diverse selection strategies to enhance predictive performance in high-dimensional biological data, such as salivary transcriptomic and miRNA profiles for gastric cancer detection.

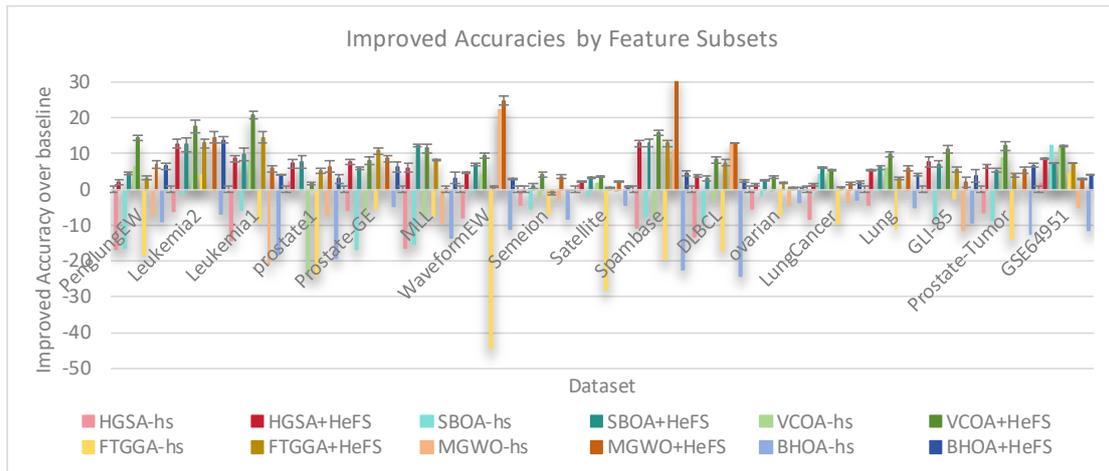

(a)

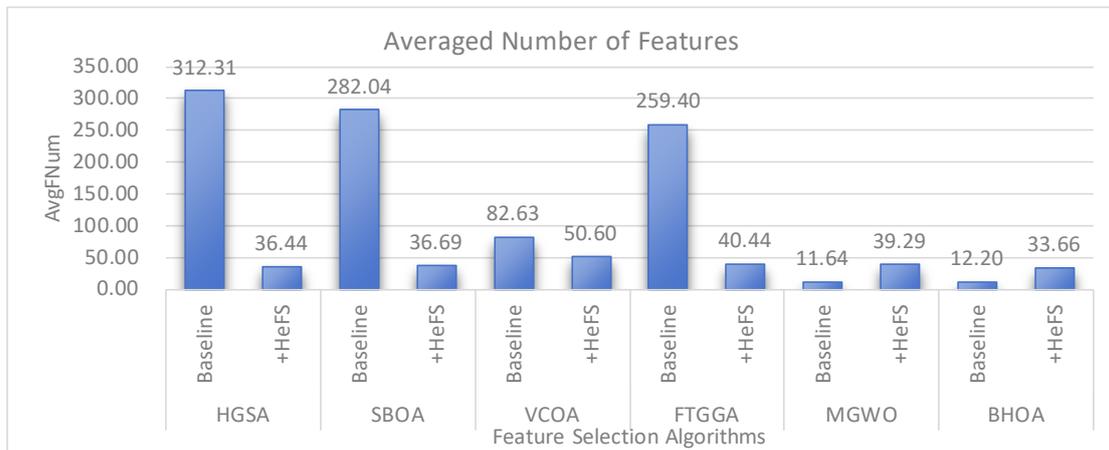

(b)

Figure 5. Comparison with the SOTA GA-based feature selection algorithms. (a) Classification accuracy (%) of helper-only variants, and combined HeFS-augmented methods across 16 datasets (standard deviation as error bars). (b) Number of features selected by SOTA methods compared to their HeFS-enhanced counterparts.

Figure 5 summarizes the number of selected features for each method. Four of the HeFS-enhanced methods (HGSA+HeFS, SBOA+HeFS, VCOA+HeFS, and FTGGA+HeFS) select fewer features than their original counterparts, while the remaining variants show only marginal increases. Even when more features are retained, the performance gains remain significant, underscoring the quality and complementarity of the helper-selected features.

Supplementary Figure S4 illustrates accuracy convergence over 100 iterations. The dotted lines represent the baseline SOTA methods. HeFS shows rapid improvements in early iterations and continues to progress steadily, ultimately surpassing the baselines by a clear margin. These trajectories confirm both the efficiency of the search process and the stability of convergence.

Supplementary Figure S5 presents boxplots of accuracy improvements achieved by HeFS over the six SOTA baselines. In almost all cases, the median improvement is positive and well above zero, while the interquartile ranges are narrow. This indicates that the helper-selected features provide consistent and robust benefits across datasets and runs, complementing conventional selections by capturing additional discriminative patterns that baseline methods miss.

**5.5 Case Study on a Molecular Property Prediction Task**

To further demonstrate the applicability of HeFS, we conducted a case study on the Toxicity dataset (Gul, et al., 2021). The evaluation was performed using a Logistic Regression (LR) classifier on three distinct feature subsets: 1) the Ttest features are the top-ranked features selected by the univariate t-test, 2) the Ttest+HeFS features are the Ttest features augmented with helper genes identified by HeFS, and 3) the Decision Tree Classifier (DTC) features are 13 features selected by the DTC method (Gul, et al., 2021).

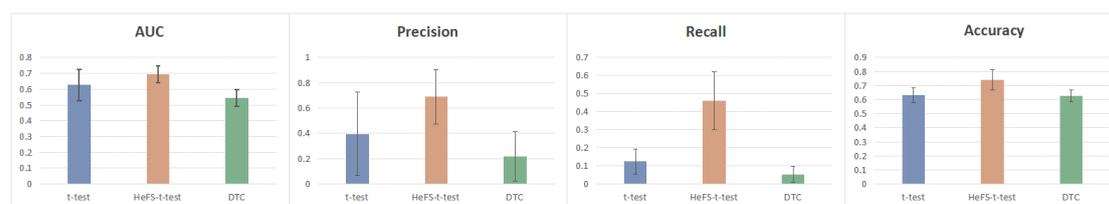

Figure 6. Toxicity prediction performance of different feature subsets. The three feature subsets Ttest, Ttest+HeFS, and DTC are evaluated for their LR-based prediction performance on the Toxicity dataset using the metrics AUC, Accuracy, Precision, and Accuracy.

Figure 6 compares classification performance across four metrics. The results show that the Ttest+HeFS features subset consistently outperforms both the Ttest features alone and the DTC-selected subset (Gul, et al., 2021). Notably, augmenting the Ttest features with helper features yields clear improvements across all four metrics, while the DTC-selected features lag behind.

We also examined the statistical properties of these features by comparing their p-values between positive and negative samples (Supplementary Figure S6). As expected, all t-test features had p-values below 0.05, confirming statistical significance. In contrast, several DTC-selected features exhibited p-values above 0.05. Interestingly, some helper features introduced by HeFS also had p-values greater than 0.05, yet their inclusion led to improved predictive performance.

This observation indicates that p-value alone is not sufficient to assess feature utility in complex, high-dimensional molecular datasets. Features deemed insignificant in univariate statistical tests may still capture nonlinear dependencies, higher-order interactions, or complementary information that enhances overall classification performance when combined with core features. This highlights the strength of HeFS in identifying valuable features that conventional selection criteria would typically discard.

### 5.6 Analyzing the Effectiveness of the Helper Set

To better understand the relationship between the CoreSet and HelperSet, we analyzed their internal and cross-set correlations using Pearson correlation coefficients (PCCs). As shown in Figure 7 (a), features within each set exhibit moderate internal correlation, whereas the average correlation between CoreSet and HelperSet features remains low (typically between -0.30 and 0.30). This low inter-set correlation indicates that HelperSet features provide complementary and non-redundant information, thereby enriching the feature space and reducing redundancy. Such complementarity is crucial for enhancing robustness and improving generalization in high-dimensional tasks.

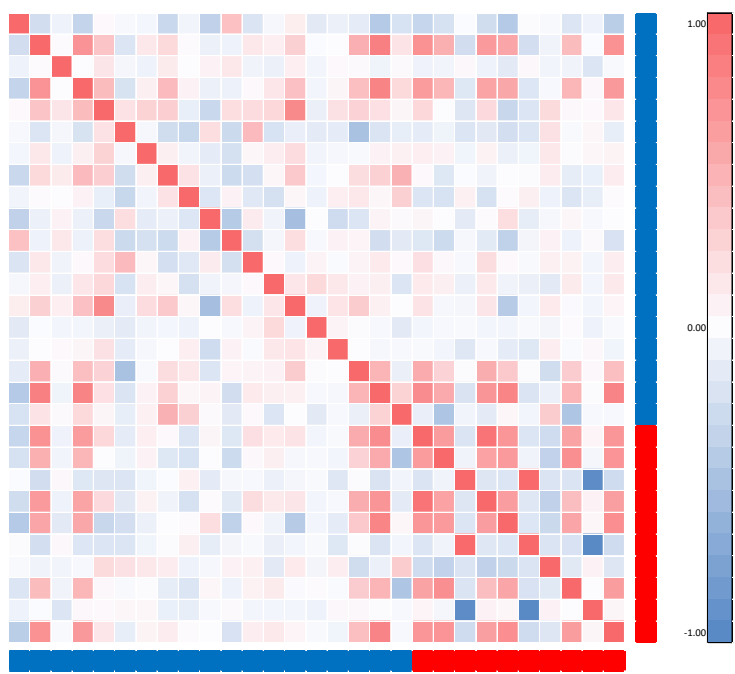

(a)

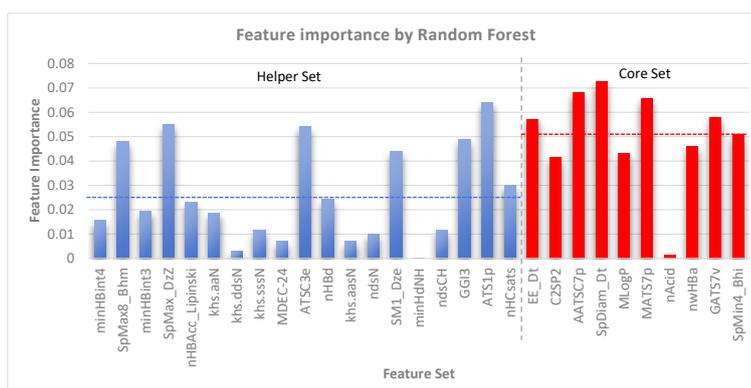

(b)                  (c)

Figure 7. Contribution evaluations of the features in the CoreSet and HelperSet. (a) PCC heatmap between features in CoreSet (red) and HelperSet (blue). Red in the heatmap denotes high positive correlation, and blue denotes negative correlation. (b) lists the feature names in the helper and core sets. (c) Feature importance scores of CoreSet and HelperSet as computed by Random Forest. Dashed horizontal lines represent the mean importance of each group.

To further assess the contribution of these features, we computed feature importance scores using a Random Forest classifier (Figure 7 (b)). The results show a clear separation in mean importance: CoreSet features generally exhibit higher importance values, reflecting their strong discriminative power, while HelperSet features have lower but non-trivial importance. Importantly, the HelperSet features contribute additional predictive signals that are not captured by the CoreSet alone. The group-wise averages, marked by red (CoreSet) and blue (HelperSet) dashed lines, confirm that although HelperSet features are individually weaker, they provide meaningful complementary effects when combined with the CoreSet.

In summary, the low cross-set correlation and distinct but complementary importance profiles validate the role of the HelperSet in enhancing the overall feature pool. By supplying diverse and less redundant signals, HelperSet features strengthen model robustness and improve classification performance beyond what is achievable with CoreSet features alone.

**5.7 Pareto Front Analysis**

To better understand the trade-offs in the multi-objective optimization process, we analyzed the Pareto fronts produced by HeFS. This analysis reveals how candidate solutions balance the competing goals of classification accuracy and feature complementarity, and highlights the role of the HelperSet in achieving effective compromises.

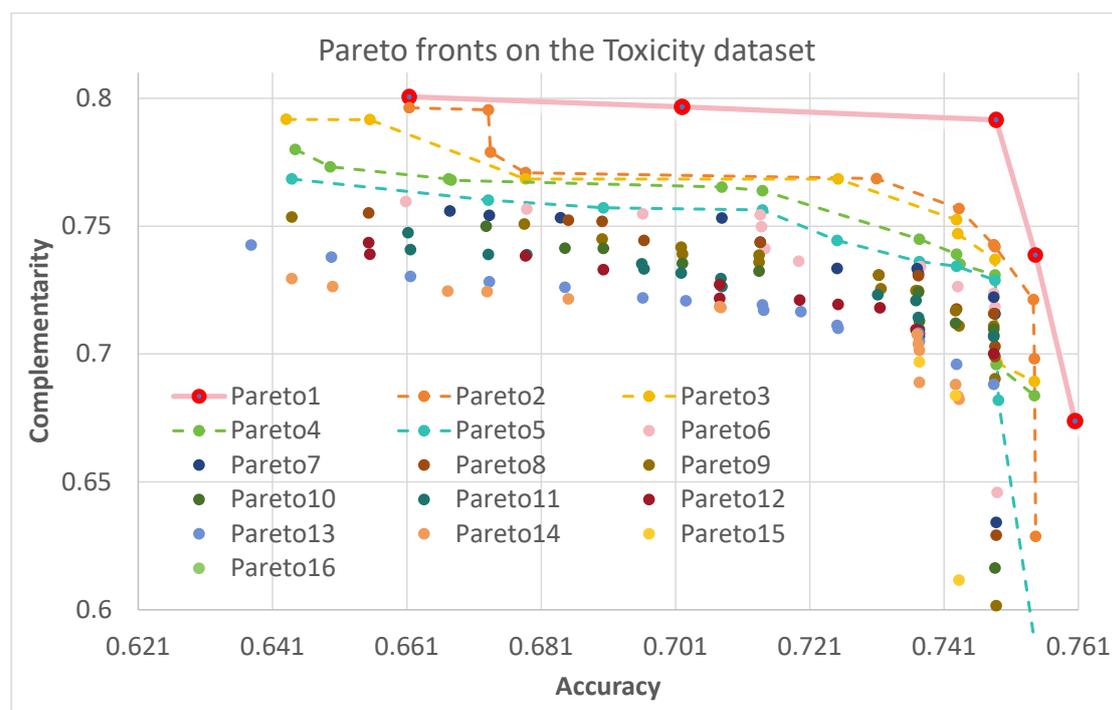

(a)

|           | nT9Ring | ndS | SsNH2 | nHBd | ndsN | nHCsats | khs.sssN | khs.aaN | khs.aasN | ATSC3e |
|-----------|---------|-----|-------|------|------|---------|----------|---------|----------|--------|
| Solution 1 | 1 | 1 | 1 | 1 | 1 | 1 | 1 | 1 | 1 | 1 |
| Solution 2 | 1 | 1 | 1 | 1 | 1 | 0 | 0 | 0 | 0 | 0 |
| Solution 3 | 1 | 1 | 1 | 1 | 1 | 1 | 1 | 1 | 1 | 1 |
| Solution 4 | 1 | 1 | 1 | 0 | 0 | 1 | 0 | 0 | 0 | 0 |
| Solution 5 | 0 | 0 | 0 | 1 | 1 | 1 | 1 | 1 | 1 | 1 |

(b)

Figure 8. Investigation of Pareto fronts on the Toxicity dataset. (a) Pareto front distribution of candidate solutions on the Toxicity dataset. The x-axis represents accuracy, and the y-axis represents complementarity. Larger markers indicate first-front solutions. (b) The patterns of the top 10 frequently-used helper features appearing in Pareto-optimal solutions. HelperSet features are prominently represented, confirming their consistent utility.

Figure 8 (a) depicts the distribution of Pareto fronts on the Toxicity dataset. The horizontal axis denotes accuracy, and the vertical axis represents complementarity. Each point corresponds to a candidate solution, with marker size and color indicating Pareto front rank. Solutions on the first Pareto front (larger markers) achieve the most favorable trade-offs, as they are non-dominated with respect to both objectives. Higher-level fronts show progressively lower quality, reflected in reduced accuracy and complementarity.

This analysis underscores the advantage of multi-objective optimization: unlike single-objective approaches, it provides a diverse set of Pareto-optimal solutions that reflect different trade-off balances. This diversity allows practitioners to select solutions that best match specific needs while improving robustness to uncertainty and complexity. By explicitly modeling accuracy-complementarity trade-offs, the optimization process is more likely to uncover globally competitive feature subsets in high-dimensional spaces.

We further examined the frequency of features appearing in Pareto-optimal solutions. As shown in Figure 8 (b), the heatmap highlights the 10 most frequently selected features across all Pareto-optimal solutions. Strikingly, 7 of these features originate from the HelperSet identified by HeFS. This strong overlap suggests that helper features are not only relevant but consistently contribute to solutions achieving balanced trade-offs. Since Pareto-optimal solutions inherently capture robust compromises across objectives, their alignment with HeFS-selected features indicates that the method effectively captures the structural dependencies in the data and directs the search toward globally strong feature subsets.

## 6. Conclusion

In this study, we introduced Helper-Enhanced Feature Selection (HeFS), a novel framework that enhances existing feature selection methods by discovering complementary features from the unselected feature space. The HeFS method integrates intelligent initialization, ratio-guided mutation, and Pareto-based multi-objective optimization to jointly optimize classification accuracy and feature complementarity.

Our framework is model-agnostic, meaning it can be applied to feature subsets selected by various methods. Extensive experiments on 17 benchmark classification datasets and molecular discovery tasks demonstrate that HeFS consistently improves performance across a broad range of traditional (e.g., RF, Lasso) and state-of-the-art (e.g., HGSA, FTGGA) GA-based feature selection methods. HeFS not only achieves higher classification accuracy but also selects fewer features, exhibiting strong generalization across high-dimensional domains. The helper features identified by HeFS are shown to provide valuable complementary information that is often overlooked by conventional methods.

Looking ahead, we aim to extend the HeFS framework to dynamic settings, such as streaming or time-series data, and explore its integration with end-to-end learning pipelines for real-time applications. We believe this work paves the way for modular and adaptive feature selection strategies in high-dimensional machine learning tasks.

## Acknowledgements

This work was supported by the National Natural Science Foundation of China (No. 82574200), the Changzhou Sci&Tech Program (to Z.W., Grant No. CJ20243019), the Scientific and Technological Research Project of the Department of Education of Jilin

Province (Grant No. JJKH20250121BS), Guizhou Provincial Science and Technology Projects (ZK2023-297), the Science and Technology Foundation of Health Commission of Guizhou Province (gzwkj2023-565), and the Fundamental Research Funds for the Central Universities (JLU).